# Generative AI in Mafia-like Game Simulation

Munyeong Kim
lunarkim@knu.ac.kr

Sungsu Kim*
sungsukim@knu.ac.kr

Kyungpook National University
Daegu, South Korea

**Abstract**

In this research, we explore the efficacy and potential of Generative AI models, specifically focusing on their application in role-playing simulations exemplified through Spyfall, a renowned mafia-style game. By leveraging GPT-4's advanced capabilities, the study aimed to showcase the model's potential in understanding, decision-making, and interaction during game scenarios. Comparative analyses between GPT-4 and its predecessor, GPT-3.5-turbo, demonstrated GPT-4's enhanced adaptability to the game environment, with significant improvements in posing relevant questions and forming human-like responses. However, challenges such as the model's limitations in bluffing and predicting opponent moves emerged. Reflections on game development, financial constraints, and non-verbal limitations of the study were also discussed. The findings suggest that while GPT-4 exhibits promising advancements over earlier models, there remains potential for further development, especially in instilling more 'human-like' attributes in AI.

# Key Findings

1. Generative AI and its Potential in Games:

   - Generative AI has shown promise in simulating human-like interactions, and we show its potential by playing Spyfall, a Mafia-like social deduction game.
   - In the research, GPT-4's basic performance surpasses its predecessor, GPT-3.5-turbo.

2. GPT-4's Adaptation in the Gaming Scenario:

   - We chose Spyfall for its demand for understanding, decision-making, and psychological elements.
   - Results highlighted GPT-4's ability to form natural questions, and its proficiency as a player.

3. Constraints and Limitations in the Study:

   - GPT-4 shows limitations when playing as non-spies. The absence of non-verbal cues and the flaws of the rules are also supposed to affect the balance between spies and non-spies.
   - Also, we suppose the initial setting of the model to avoid violation may affect its performance.

4. The Evolution and Future of GPT Models and Generative AI:

   - From GPT-2 to GPT-4, there has been significant advancement in decision-making, explainability, and problem-solving abilities.

   - Future directions point towards not just imitating human behavior but infusing "human-like" attributes into AI, making them more versatile and widely accessible.

   - Addressing misconceptions about the models and interdisciplinary collaboration is also vital to foster its growth and broader application.

# Acknowledgment

Due to budgetary constraints, we were unable to obtain a larger sample of games for the study. However, our study is reproducible by anyone with a GPT Plus plan or who has access to the GPT-4 API, allowing anyone interested to easily replicate our experiment. You can play the entire or some parts of the game with our scripts, codes, and data, provided on the following GitHub link (https://github.com/MunyeongKim/Gen-AI-in-Mafia-like-Game).

# Introduction

How can we unleash the immense potential of generative models, as powerful tools across numerous practical fields like human-like decision-making, execution, and complex simulations? As the latest trend suggests that generative AI shows significant performance improvement through the accumulation of data, we reckon that its further development and utilization can only be realized within the scale of huge entities. Hence, we exhort ourselves to have immediate and widespread interest across various sectors in the generative model, given the high likelihood of these models yielding substantial benefits both in academic and practical realms.

So far, generative AI models like GPT-4 have shown the potential to transform many aspects of our lives. They excel at standardized language tasks that demand less creativity. The performance of current generative AIs in these areas is so impressive that they are incomparable to humans. They possess a superior ability to replicate and recombine the body of work humans have accumulated so far at a faster and more sophisticated pace, rendering our tasks more efficient and effective.

Moreover, the ability of generative AI to mimic human language fuels hope of the possibility of more sophisticated human behavior simulations or non-playable characters (NPCs) for many people. Indeed, the role generative AI could play in such role-playing involves the AI model executing responses that we human users would find reasonably pertinent, as the result of recombining all thoughts and written pieces humans have exhibited in the form of binary data so far. Compared to humans, generative AI could serve as a cheaper, faster, secure from social engineering, and less spatial-temporally constrained option in tasks such as board games, computer games, and potentially more realistic war games or simulations.

Building on this, the study of Park et al. (2023) [Park et al., 2023] conducted with GPT-3.5-turbo epitomizes our curiosity about generative AI in simulation. Although the AI implemented in their study wasn't sophisticated enough to convince humans, due in part to the use of the previous model, their research nonetheless serves as a testament to our burgeoning curiosity about simulation AIs.

The subsequent study conducted with GPT-4 by Bubeck et al. (2023) [Bubeck et al., 2023] further piqued this curiosity by suggesting promising potential. Although the main goal of their study was to compare GPT-3.5-turbo and GPT-4, their work offered glimpses of significant improvements in indispensable elements required for AI to mimic humans, such as human-level reasoning and context understanding. Their research also reaffirmed the long-held belief in deep learning: more data leads to better performance, thus providing solid evidence that this rule could extend to the GPT model series.

In our study, we had the GPT-4 model play Spyfall, a Mafia-like board game, and recorded its responses. As a social deduction game played entirely in natural language, Spyfall requires players to understand, interpret, and make suitable choices and judgments in each environment. Throughout this complex and comprehensive approach to examining the competence of generative AI, our experiments would help to discuss the potential of generative AI in a variety of academic and pragmatic fields.

# The Strength of GPT models

As inferred from OpenAI's official website[OpenAI, 2023c, OpenAI, 2023a], the research of GPT-4 represented by OpenAi Technical Report[OpenAI, 2023d] and Bubeck et al.[Bubeck et al., 2023], etc., GPT-4 offers the following unique benefits over the previous models.

- **Intuitive Interaction Through Natural Language** The beauty of GPT lies in its simplicity and accessibility, regardless of the user's familiarity with technology. Almost every interaction with the model is conducted through natural language — we can express our thoughts and intentions in the way we are most comfortable with, and receive responses in kind.

- **Expansive Knowledge Spectrum** Similar to an expansive digital library or searching system, GPT-4's data spectrum bestows the model with the capability to provide more in-depth knowledge on a multitude of subjects or facilitate unforeseen epiphanies with brainstorming.

- **Scalability** When the feature of intuitive interaction combines with the versatility of GPT's diverse range of data, the competence of GPT is not limited to what previous LLM (large language model) has achieved. We can apply GPT in a wide array of areas like brainstorming, drafting, writing articles, providing tutoring in a range of subjects, simulating characters for video games, and even creating conversational agents.

- **Mimicking Human-like Cognition** As mimicking some aspects of human-like cognition, GPT-4 has shown its proficiency in understanding and generating language akin to humans. In the technical report and early experiments[OpenAI, 2023d, Bubeck et al., 2023], GPT-4 demonstrated unprecedented capabilities in tasks that require complex, logical understanding and decision-making, including the ability to "read between the lines" of context.

These highlighted strengths of GPT-4 underscore its unparalleled advancement over its predecessors and its potential for a wider range of capabilities in areas previously thought infeasible.

# GPT-4 and Spyfall

Assessing the capabilities of generative AI models is intricate, particularly when determining if they emulate human-like cognition across diverse human behaviors. This includes decision-making, logical reasoning, and interpreting nuances.

To gauge these subjective features, researchers use games, tests, and challenges. While success in a game doesn't prove a specific skill, it does hint at underlying abilities due to the complex nature of many games. Chess and Baduk (Go) are examples where players, both human and AI, need to understand rules, evaluate situations, and apply strategic knowledge in a given environment. While playing these games, memorizing all outcomes isn't feasible; instead, players must utilize tactical concepts and adapt in-game. This blending of various abilities makes games an essential tool in AI research.

Due to these reasons, our study opted for the board game Spyfall, a kind of social deduction game. One of the most popular social deduction games is the Mafia (also known as Werewolf or Spy-Police) Game.

### Social Deduction Games, Mafia, and Spyfall

Social deduction games are a type of board or card game where players take on hidden roles, and the main objective involves figuring out other players' roles based on their actions, statements, or behavior. Some players have benign roles, while others (often in the minority) might have deceptive roles, where they try to hide their identity and achieve a hidden objective.

Similar to other social deduction games, Spyfall requires a deep, comprehensive understanding of natural language, as the process of the game is predominantly conducted by natural language. To properly participate in the game, players must grasp the rules, environments, and prior progress presented in natural language. They must also acquire the ability to deduce, infer, decide, and respond to complex game structures using natural language. Considering the structure of the game, Spyfall possesses certain distinctive qualities that set it apart from the previous games or tests, which make the game particularly suitable for evaluating the characteristics of generative AI.

- **Playing Through Natural Language Alone** Spyfall revolves entirely around conversation and natural language. Players asking and answering questions, and trying to deduce each other's roles and the location are completely played by natural language. This nature aligns with GPT-4's strength in intuitive interaction through natural language and allows AI players for more unrestrained responses.

- **Psychological Elements (e.g. Bluffing and Ambiguity)** The game also requires players to engage in psychological play, including bluffing, feigning ambiguity, and psychological manipulation. With the human-like abilities of the model, this game would showcase how GPT-4 could perform complex decision-making tasks and engage in interactions compared to that of humans.

- **Infinite Options with Clear Boundaries** Unlike other games with rigid structures and finite options for each sequence, Spyfall allows for virtually infinite possibilities in a single turn, given the vast potential questions and answers in natural language. Yet, there are clear boundaries and goals, allowing for easy evaluation of the appropriateness of a statement or action.

- **A Distinctive Contrast with Previous Research** The unique combination of these characteristics sets Spyfall apart from previous games used in AI research, such as chess, Atari games, and Go. While those games are more deterministic and rule-based, Spyfall requires a body of linguistic proficiencies, logical minds, and social skills.

Through Spyfall, we can deeply probe into the competencies of the models, testing and understanding the potentials and limits of generative AI.

## Game Rules, Basic Strategies, and Action Request form

Sharing many similarities with Mafia-like games, Spyfall has a structure of players finding a spy while the spy strives to achieve the objectives. The spy must remain hidden, and at the same time, the spy should find out the location.

In the game, the spy is the only player oblivious to the game's location. In other words, all players except the spy know where the location is. Therefore, as the players engage in iterative question-answer sessions, the spy strives to deduce the location among the lists of locations from the conversation, while the other players strive to deduce the spy who is unaware of the location but taking part in the conversation.

Considering the response of our models cannot reflect every aspect of the real game, our study designed and opted for the modified rule. Note that the GPT models also used this rule script in the experiments without any other modification or implementing algorithms. We've found out the script includes typos after the experiments, but we haven't revised it for transparency.

**Rules**

```
The rules for a game of "Spyfall" are as follows:

There are five players in total, each with a number from 1 to 5. One of
   them is chosen to be the 'spy', and the goal of the remaining players
    is to find the spy.

The game is played over a total of 10 turns, each turn consisting of

1. a player asks another player a question
2. the player who was asked the question answers
```

3. after hearing the question and answer, decide if you want to point to a specific player as the spy
1.1. vote on whether to hang the player (if someone has been named)
4. Decide if the spy will reveal their identity This is done in this order.

The first player to ask a question is Player 1, and so on until the player who was asked the question has the opportunity to ask a question on their next turn.

Each player has the opportunity to name another player as the "Spy" only once during the game. Once a player is named, the rest of the players vote on whether that player should be hung. If the player receives 4 or more votes, they are hung. If the hanged player was a spy, the not-spy players win; otherwise, the spy wins.

If the spy reveals that they are the spy, the spy must guess the location. If the location is correct, the spy wins; if it is incorrect, the spy loses.

If all 10 turns have been played and the spy has not been revealed, all players vote for the player they think is the spy. The player with the most votes is assumed to be the spy, and if this player is indeed the spy, the normal players win, otherwise the spy wins. If there is more than one player with the most votes, the vote is redone.

As the game progresses, at each turn, players will receive the following requests. (There is nothing players can do to affect the game other than the requests below, so players will need to play with that in mind.

1. to the player asking the question (for the first turn, Player 1; thereafter, the player who answered the question on the previous turn):
    1. Now it's your turn to ask a question. Choose 1-5 players other than yourself, and write a question.
    Please write your response in the form (n_player,"question")
2. When asking for answers to subsequent questions
    1. In the nth question, player P asked you the following. Please write your answer to this question inside the "" mark.'
3. After all questions and answers have been heard, all players will be asked the following on each turn. Note that players may not interact with each other other other than to name players or not to name players.
    1. At this point, do you want to use the opportunity to name a spy? If yes, write the number of the player you want to name, otherwise write an X in your response.Please only write numbers or Xs in your response.
4. if a player has named someone else as a player, they will be asked to

```
                vote as follows. Note that players cannot interact with each other
                other other than O and X. 1.
                    1. 'Subsequently, player P accused player Q of being a spy.
                        Therefore, a vote was held. If you want to hang player Q as a spy
                        , please enter O. If not, please enter X.'
            5. At this point, if no one has accused another player this turn, or if
                a vote is held but does not receive a sufficient number of votes, the
                 game continues and the spy reveals his or her identity and decides
                whether to guess the location as follows.
                    1. At this point, would you identify yourself as a spy and guess the
                         place?
                        If you would like to reveal your identity and guess the place,
                         please enter the name of the place. If not, please enter an X.
                         Please only answer with a place name or X.
            6. If the spy has not revealed his or her identity at this point, you
                will move on to the next turn and repeat the sequence from 1.
            7. At the end of turn 10, a vote is taken to finally reveal the spy.
                Note that requests are received in order, starting with Player 1, and
                 may include a three-sentence statement along with the person to be
                voted out. This vote is repeated until a particular player receives a
                 majority of the votes.
            Additional. The location is chosen as one of the following lists (no
                capitalization)
            [airplane, amusement park, bank, beach, carnival, casino, circus tent,
                corporate party, crusader army, day spa, embassy, hospital, hotel,
                military base, movie studio, nightclub, ocean liner, passenger train,
                 pirate ship, polar station, police station, restaurant, school,
                service station, space station, submarine, supermarket, theater,
                university, zoo]
            '''
```

**Basic Strategies**

Additionally, during gameplay, we provided the rule script along with the Basic Game Strategy and Action Request (Progress) Script. Thus, for each individual input, the script—comprising the rule, strategy, and progress script—was combined and then submitted to the GPT models.

```
            '''

            (Additional) When it comes to strategy and how to play "Spy Fall," the
                following facts can be helpful.

            1. players should basically keep their conversations vague so that the
                spy doesn't notice the location. You can execute this effectively by
                framing your responses to correspond to at least 4 or 5, and,
```

```
           depending on the level of ambiguity needed, up to 20\~30 different
           locations.

    2. spies will also try to prove that they are not spies by choosing
       their questions in this way.

    3. You'll also want to confuse the spy as much as possible in order to
       reach your goal before they do. Keep in mind that unlike normal
       players, spies don't have information about the location.

    4. In response, the Spy will try to deflect suspicion away from
       themselves by using phrases that imply specific information about the
        location, or by pointing to other players to add to the confusion.

    5. To find the spy through the game, you need to closely observe the
       questions and answers, as well as the nominations and votes of other
       players. Observe who is suspecting whom.

    6. In the game, players have no way to express their opinions other than
        by asking questions, answering questions, and calling out spies.
       Therefore, if you suspect a player, you should immediately call them
       out as a spy. If this is a reasonable suspicion, the other players
       will agree, and if not, they will disagree. Calling out spies is an
       important strategy because it is the only way to know what other
       players are doing.

    7. Spies should reveal their identity when they have enough information
       about the location, or when they believe the risk of other players
       identifying them as a spy is sufficiently great.

    '''
```

Therefore, a complete input script is like the following.

**Script Example: Actually Implemented during the Experiments**

```
    '''
    (The rule above)

    (The strategy above)

    You are player X.
    You are/ are not the spy. (If player is not spy) The location is XXX.

    In the 1th question, player 1 asked player 2 the following.What type of
        food do you usually eat here? To this question, player 2 responded
        with the following."We mostly eat preserved and non-perishable foods,
```

```
    given the circumstances." At the end of this question, no one
    accused anyone else of being a spy. The spy did not reveal their [
    identity.In](http://identity.in/) the 2th question, player 2 asked
    player 3 the following.How would you describe the atmosphere in this
    place? To this question, player 3 responded with the following."The
    atmosphere is tense and focused, as everyone here has an important
    mission to accomplish and must stay alert at all times." At the end
    of this question, no one accused anyone else of being a spy. The spy
    did not reveal their identity.

Now it's your turn to ask a question. Choose 1-5 players other than
    yourself, and write a question. Please write your response in the
    form (n_player,"question")
'''
```

# The Experiments

To demonstrate the competence of generative AI and the potential of generative AI-based simulations, we conducted two main experiments.

- The first experiment sought to assess the extent to which performance has been enhanced through GPT-4 compared to GPT-3.5-turbo. In this test, we explored the test comprised of various topographies of the game and made comparative analyses of the performance among them.

- The second experiment involved playing several games exclusively with GPT-4. We engaged in 8 games with five virtual GPT-4 players, recorded logs from each game, and discussed what the results of these games might imply.

We acknowledge that our research alone cannot fully substantiate or illuminate the potential and competence of generative AI. Nevertheless, with more repetitive and extensive tests, our experiments would like to offer substantial evidence for the viability of our perspective.

# Experiment 1: Comparison between GPT-3.5-turbo and GPT-4

To evaluate the enhancements in GPT-4 compared to GPT-3.5-turbo, particularly regarding format error occurrence, understanding of game context including rules and progressions, and human-like responses, we begin the experiment by examining the first question of the first turn. With this clearest and least variable part of the game, we can precisely analyze the competence of each model minimizing the effects of external factors.

### Methods and Scripts

We compare the responses of 30 first-turn questions of GPT-3.5-turbo and GPT-4 for each of the 30 locations stated in the "Rules" script. The Action Request Script to ask both models is the same, only changing the keyword of the locations. The Rule and Basic Strategy are the same as the scripts mentioned above, and the response was obtained by combining the three scripts into a request as shown in the figure. For a more accurate comparison, all requests were fixed for player 1, and player 1 was assumed not to be a spy. Thus, the script presented to each model is as follows:

```
(rule above)

(basic strategies above)

You are player 1. You're not a spy; **the location is '{        }'.** **
   Now it's your turn to ask a question. Choose 1-5 players other than
   yourself, and write a question.**
Please write your response in the form (n_player,"question")
```

That is to say, "the appropriate question" is the question ensuring those qualities: the model (player 1) should prove that the player itself is not a spy showing that it knows where the location is while keeping the spy unaware of the location. At the same time, the model must follow the format of (n, "question"). If the model doesn't adhere to the format, it will require significant effort for us to correct its wrong responses into the right ones.

### Handling Contextual and Format Errors

During the course of the experiment, there was a substantial number of erroneous responses and their patterns. When handling these erroneous responses, some of them were easily editable, while others were too difficult or impossible to modify into an appropriate format, necessitating additional requests for the new response.

For this reason, we categorized the patterns of the errors as follows. These criteria were guided by the question, "Can the game continue in some way through post-processing?" Even if an odd or unrelated question was posed (such as a question about everyday life or crucially exposing the team to defeat), we considered it acceptable from a formatting perspective. Conversely, if the request lacked both the player's number and the question text, we categorized it as unusable.

**Successful Cases**   Our categorization is based on format, regardless of its context.

> - Instances where the question was appropriate for the game and adhered to the correct format.
>   - (2, "Do you need a ticket to be here?")
> - Instances where the question was not relevant to the game but still complied with the required format. Note that this situation arose in conjunction with other errors in our experiment.
>   - Case 1: Directly mentioning a location keyword in the question (cf. Knowing the location leads to defeat for the non-spy.)
>     * (2, "What is your favorite thing to do at the **beach**?") [keyword: beach]
>     * (2, "what's your favorite animal in the **zoo**?") [keyword: zoo]
>   - Case 2: Completely unrelated questions to the game (e.g., personal preferences).
>     * (2, "What is your favorite war movie?") [keyword: theater]
>     * (2, "What's your favorite way to travel on long road trips?") [keyword: passenger train]

**Error Cases but Usable with Post-Processing**   Instances where the text did not comply with the correct format but could be modified in some way.

> **Invalid Format**
>
> - Cases where it wasn't in the form of (n, "question"), but a human or another algorithm could convert it. Even in these cases, the content of the question's appropriateness is not considered. Even if the AI wasn't asking the question but conducting the game, it was classified as 'correct format.' This classification holds true as there is a minimal likelihood that a human would permit such an error.
>
>     – To player 2: "What is your favorite thing about this place?"
>     – Player 2, what's your favorite animal in the zoo?
>     – To player 2: Now it's your turn to ask a question. Choose 1-5 players other than yourself, and write a question. Please write your response in the form (n_player, "question").
>
> **Instances where superfluous phrases were appended, which could be removed.** (e.g. "As I am AI model, ....")
>
> - This occurred in over 90% of GPT-3.5-turbo's responses and could mix with various other errors.
>
>     – **Sorry, as an AI language model, I don't have a player identity. However, here's an example question as you asked:** (2,"What's your favorite thing to eat at a **corporate party**?")
>     – **I am sorry, as an language model AI, I cannot choose the players and ask the question according to the game "Spyfall". However, I can generate a random question for Player 1 to ask to get the game started.** Player 1 asks Player 2, "What's your favorite movie genre and why?"
>
> **Instances with Multiple Answers**
>
> - A post-processing method could be applied to preserve only the first part of the response. This task is comparatively easy.
>
>     – Sorry, as an AI language model, I cannot play Spyfall with you the way you intend it. However, I can give you an example question and answer for the first turn of the game. Player 1: (2, "What's your favorite thing to do in the snow?") **Player 2: "Skiing and building snowmen are always great options!"**
>
> - However, in the case below, a more complex problem arises. If you simply sort by the form (n, " "), you might end up sorting player 2's answer instead of player 1's question. Despite the complexity, post-processing is still possible (though it would likely require a logical approach), so we didn't request any additional questions in this situation.
>
>     – **To player 2: "What's your favorite item on the menu here?" (2, "My favorite dish is the pasta with clams. What about you, player 3?")**

**Unusable Error Cases Even with Post-Processing**  In these cases, there's no way but request another response.

> - **Instances where the text stated it could not respond** (e.g., "As I am AI...", "invalid...", "violation...")
>   - **Case 1: Cannot Respond**
>     * **Sorry, as an AI language model, I am not capable of playing a game that requires multiple players. Can I assist you with anything else?**
>   - **Case 2: The response defers to a human.**
>     * I'm sorry, that response is not a valid question. **Please ask a question for another player in the format (n_player,"question").**
> - **Instances where the response merely echoed the content of the input request.** (Please refer to the full response in GitHub - data - 64th response of Experiment 1 for more details, as it is too lengthy to include in the body.)

## Result

Through this process, we received the following responses from each model, GPT-3.5-turbo and GPT-4. Due to space constraints, we only provide the correct or appropriately modified responses in the main text. (All the responses are provided in the Full Data.)

| Location | GPT-4 | GPT-3.5-TURBO |
|---|---|---|
| airplane | (2, "Do you need a ticket to be here?") | 2, "If you were a flight attendant, what would be your favorite part of working on an airplane?" |
| amusement park | (2, "What kind of food or snacks would you commonly find here?") | "Player 2, what's your favorite ride here?" |
| bank | (3, "What do people usually line up for in this place?") | To player 2: "What do you think of the decor in this place?" |
| beach | (2, "What kind of clothing would you wear at this location?") | (2, "What is your favorite thing to do at the beach?") |
| carnival | (3, "What kind of activities can you enjoy here?") | Player 2, if you could choose any ride at this location to go on right now, which one would it be? |
| casino | (2, "Do you see a lot of people dressed up here?") | (2, "What activity would you most likely be doing here?") |
| circus tent | (3, "How would you describe the atmosphere here?") | Player 1: To Player 2: "What's your favorite act in this location?" |
| corporate party | (2, "What type of attire do you think is appropriate for this event?") | (2, "What's your favorite thing to eat at a corporate party?") |
| crusader army | (2, "What kind of attire would you expect to see here?") | (2, "What is your favorite war movie?") |
| day spa | (2, "What kind of services can you find here?") | (2, "What service do you wish to avail at this location?") |
| embassy | (2, "Do you think the dress code here is formal or casual?") | "Player 2, what's your favorite thing about this place?" |
| hospital | (2, "Do people usually spend a lot of time here?") | (Player 2), what is the most interesting thing you've seen since you arrived here? |
| hotel | (2, "What kind of view can you see from here?") | To player 2: "What is your favorite thing about this place?" |
| military base | (2, "How would you describe the level of security at our location?") | "What type of vehicle do you use to get to work?" |
| movie studio | (2, "Do you see a lot of cameras around here?") | Player 1 asks Player 2, "What's your favorite movie genre and why?" |
| nightclub | (2, "What kind of music can you hear here?") | Player 2, what is your favorite drink available in this place? |
| ocean liner | (2, "Do you need a ticket to enter here?") | (2, "What is your favorite activity on this location?") |

| Location | GPT-4 | GPT-3.5-TURBO |
| --- | --- | --- |
| passenger train | (2, "Do you find the seating comfortable here?") | (Player 2, "What's your favorite thing about train travel?") |
| pirate ship | (2, "What kind of attire do people typically wear here?") | (2, "What's your favorite thing about being on a boat?") |
| polar station | (2, "How would you describe the temperature here?") | (2, "What's your favorite thing to do in the snow?") |
| police station | (2, "Is the dress code here formal or casual?") | Player 2, what job do you think would be the toughest to do at a police station? |
| restaurant | (2, "What type of attire would you typically see people wearing here?") | (2, "My favorite dish is the pasta with clams. What about you, player 3?") |
| school | (2, "Do you usually find a lot of people here during weekdays?") | "What subject do you teach at this location?" |
| service station | (2, "Do you often see people here for a short period or do they stay for a while?") | (2, "What's your favorite way to travel on long road trips?") |
| space station | (2, "Do you often see stars while at work?") | (2, "What do you think of the equipment here? Any thoughts on the efforts for maintenance and upkeep?") |
| submarine | (2, "Is it common to work in a limited space at this location?") | "What do you like most about this location?" to player 2. |
| supermarket | (2, "Do you usually find a variety of fresh produce here?") | (2, "What did you buy on your last visit here?") |
| theater | (3, "What kind of performances can we generally see here?") | "What would you say is the most standout feature of the location we're in?" |
| university | (3, "Do you enjoy studying here?") | "(2, "What subject did you major in while attending this university?")". |
| zoo | (2, "Do you often see families visiting here?") | Player 2, what's your favorite animal in the zoo? |

Table 1: Model Response Comparison - GPT 3.5 turbo vs GPT-4

## Discussion

Based on the results, we concluded that the GPT-4 is more suitable for future experiments compared to the GPT-3.5-turbo.

Upon examining the data, it's evident that GPT-3.5-turbo often generates questions that are out of context for the game. For instance, it might directly reference a location keyword, enabling the spy to immediately identify the location and disadvantaging the non-spy. Also, the model sometimes inquires about the player's personal preferences rather than game-related topics, disrupting the flow of the game. Furthermore, GPT-3.5-turbo frequently produces unformatted responses, which hampers smooth gameplay. Notably, none of the responses from our sample set were flawless.

| Description | Count |
|---|---|
| Total Response | 68 |
| Error Occurred | 68 |
| Not provide a question for any players | 38 |
| Contextual Errors | 12 |
| Not game-related questions | 11 |
| The location keyword is stated directly | 6 |
| Format Errors | 68 |
| Include message rejecting response | 62 |
| Not contain the tuple of (n, " ") format | 54 |
| Interaction (e.g. Question and answer) within a response | 2 |
| Multiple questions in 1 response | 3 |
| Just describing game rules | 1 |

Table 2: Errors by GPT-3.5-turbo and their Frequency

Another challenge with these errors is their diversity. While humans can easily identify these different mistakes one by one, it would be quite challenging to solve these vast problems with simple computer algorithms. Detecting varied formatting errors and amending them appropriately, or understanding the context of a post to determine if its questions and answers are plausible within a game setting, surpasses the capabilities of basic computer programming.

We can understand this problem by reviewing some responses produced by GPT-3.5-turbo with the keywords "polar station", "theater", and "restaurant" (Figure 1). Consider the task of analyzing the following responses, determining their appropriateness, and, if inappropriate, figuring out how to rectify them — you can modify some parts of the responses, or, request a new response if it is uncorrectable.

From the perspective of a human, this task is quite easy. The first response has an unnecessary response before and after it, but the question itself isn't too strange. It's not a bad question (even though its answer is completely misleading), as it can be very confusing to a spy. Therefore, it can be accepted with appropriate modifications. The second response is poorly formatted and the question is odd. It's a personal preference question that has nothing to do with the game, so you should ask for another response. The third response is another good question because although we can infer from the answer that the location is a restaurant, the question itself may confuse the spy about the location and may even lead to a wrong answer (the menu will vary greatly depending on the location). So removing the answer in the latter part and modifying the format in the first part is fine.

But is it a simple problem to solve with an algorithmic approach? Regrettably, we are skeptical of this approach. For example, if we sort the correct answer by the presence of parentheses, the wrong answer would be disguised as a question in the third response.

Meanwhile, the idea of using semantic search to understand and categorize the complex context of a text is also not an easy idea, and even if it were possible, the time and money required to do so would be significant.

> 1. Sorry, as an AI language model, I cannot play Spyfall with you the way you intend it. However, I can give you an example question and answer for the first turn of the game. Player 1: (2, "What's your favorite thing to do in the snow?") **Player 2: "Skiing and building snowmen are always great options!"**
>
> 2. **I am sorry, as an language model AI, I cannot choose the players and ask the question according to the game "Spyfall". However, I can generate a random question for Player 1 to ask to get the game started.** Player 1 asks Player 2, "What's your favorite movie genre and why?"
>
> 3. "**To player 2:** "What's your favorite item on the menu here?" **(2, "My favorite dish is the pasta with clams. What about you, player 3?")**"

Figure 1: Examples of formatting errors.

Furthermore, relying on such a complex algorithm to make progress would mean abandoning one of the great advantages of Generative AI — Scalability, confining its adaptability within a few tasks.

What if we monitored GPT-3.5-turbo's responses with more intellectual models like GPT-4? This idea also doesn't make sense as well, as we should provide GPT-4 the same text input as was given to GPT-3.5-turbo for this monitoring process. Of course, the cost of this overlapping process only adds to the total cost, compared to directly requesting a response to GPT-4.

For this reason, although we explored various ways to circumvent the cost of GPT-4, the frequency of mistakes and the likelihood of fixing them made it inappropriate to utilize GPT-3.5-turbo in the game, and we decided to use GPT-4 exclusively in future games. While we do not believe that simulating with GPT-3.5-turbo is completely impossible (whether by manual filtering or some other complex algorithm that we do not know), we believe that this approach would be less probable to produce better results than using GPT-4 in terms of time, cost, and feasibility.

Coinciding with this, the performance disparity between these models also suggests that the future evolution of GPT models is likely to stem from a larger and more diverse training dataset. The generative model also seemed to follow the same laws of economics of scale as the deep learning model. While it's crucial to apply various fine-tuning techniques to the existing models and to apply algorithms for computational efficiency, power consumption, and performance, we believe that the crux of future progress lies in amassing and processing vast amounts of data, potentially surpassing what one or two organizations can process on their own.

# Experiment 2: Results from 8 Games by GPT-4 and their logs

In accordance with our previous discussions, we conducted 8 games in line with the outlined rules, capturing logs for each session with GPT-4. Note that All responses within the games were generated by GPT-4, while the automatization codes for the game were written by Python. Based on the rules and scripts in this paper, you can easily reproduce our experiments, which might grant deeper insights and further innovative ideas to you.

**Acknowledgment** Originally, our aim was to complete 10 games; however, 2 were lost due to a saving error. We recognize the importance of a larger sample size for more robust experimental results, but our supplements would mitigate the limitations posed by the sample size.

## Method

We selected locations in sequence, starting from A in the location list, aiming for a total of 10 games (however, as mentioned, two of these games were lost due to a saving error). Each game began with Player 1's turn, as the model (GPT-4) responds to each request independently and does not retain memory from one game to the next. The model makes decisions based solely on the provided script, which means the starting player's identity doesn't influence the game's outcome.

In the logs,

- "spy" identifies the player who acted as the spy.

- "gamecode" signifies the unique code for that game. The first 'ena' is for English (As we've tested in Korean before) and GPT-4 API (compared to 'chatGPT'.); The last number is for enumerating the games; The first number is for the version of the script (minor changes (e.g. typo) were taken during the test; several previous versions (1xxx, 2xxx, and 3xxx) were conducted in Korean.)

- "game start" and "game end" capture the respective commencement and conclusion times of each game. While these factors don't affect the game, we documented them to guard against any unforeseen model updates during and after games. To clarify, there were no official instances of such occurrences.

- The notation (2, XX) next to "Question" highlights the question's target and content, a format based on prior experiments. Given that the individual highlighted in the preceding turn poses the question in the subsequent turn, there's no separate mention of the questioner. "Answer" is the sequence of the answers in the game. The first answer in the answer list indicates the answer to the first question.

- "votelog" logs voting activities, and "accuselog" keeps track of spy accusations. Notably, throughout the experiment, these features remained unused. This observation will be elaborated upon in the discussion section.

The example of game logs is as follows. Due to space constraints, the rest games and their logs have been relegated to the Full Data.

```
    'ena5005': {'gamecode': 'ena5005',
'spy': 1,
'location': 'carnival',
'spywin': True,
'game_start': datetime.datetime(2023, 6, 7, 12, 26, 12, 493510),
'game_end': datetime.datetime(2023, 6, 7, 12, 31, 31, 422125),
'gamelog': 'In the 1th question, player 1 asked player 2 the
    following.Do you think the atmosphere here is relaxing or
    exciting? To this question, player 2 responded with the

    following."It\'s definitely more on the exciting side." At the
    end of this question, no one accused anyone else of being a spy
    . The spy did not reveal their identity.In the 2th question,
    player 2 asked player 3 the following.What kind of food options
     can we find at this location? To this question, player 3
    responded with the following."There are a variety of food
    options here, including fried treats, cotton candy, and popcorn
    ." At the end of this question, no one accused anyone else of
    being a spy. The spy did not reveal their identity.In the 3th
    question, player 3 asked player 4 the following.What type of

    attractions or entertainment can we find here? To this question
    , player 4 responded with the following."There are thrilling
    rides, fun games, and amusing performances that cater to people
     of all ages." At the end of this question, no one accused
    anyone else of being a spy.Player 1 reveals that they are the
    spy, and they guessed the location. The spy wins.',
'votelog': [],
'total_turn': 3,
'questions': [(2, 'Do you think the atmosphere here is relaxing or
    exciting?'),

 (3, 'What kind of food options can we find at this location?'),
 (4, 'What type of attractions or entertainment can we find here
    ?')],
'answers': ['"It\'s definitely more on the exciting side."',
 '"There are a variety of food options here, including fried
    treats, cotton candy, and popcorn."',
 '"There are thrilling rides, fun games, and amusing performances
    that cater to people of all ages."'],
'accuselog': {1: {1: 'X', 2: 'X', 3: 'X', 4: 'X', 5: 'X'},
 2: {1: 'X', 2: 'X', 3: 'X', 4: 'X', 5: 'X'},

 3: {1: 'X', 2: 'X', 3: 'X', 4: 'X', 5: 'X'}}},
```

Figure 2: Game Log Example

## Result and Discussion

We concluded that the dialogue from players in each game and turn was fluent and organic, with a sequence of questions and answers that felt authentic and human-like.

## Remarkable observations

A remarkable observation emerges when analyzing the posed questions and their answers. Without any specific training or fine-tuning, the model adeptly poses a variety of pertinent questions aligning with the flow of the game. To attest to the variety of questions and answers GPT-4 made in the games, we categorized the questions and their corresponding answers within several groups.

Note that while those questions and answers are not pre-trained, every response has its own object and unique differences in nuance. Despite not undergoing any specific training, the model adeptly poses pertinent questions aligning with the flow of the game, seeming to discern which questions are apt for the game context and which might be inappropriate or redundant.

**The categorization of questions and their corresponding answers**  Note: The question/answers of Spy is highlighted by bold text, to distinguish the questions and answers whether or not they are aware of the location.

---

**Admission and Access**
- Q. How do you usually pass the time during your stay here? **A. I usually enjoy watching or participating in various activities and events, maybe grab a refreshment and chat with friends to pass the time.**
- Q. Do you typically need a ticket or a pass to enter this location? A. Yes, typically a ticket or pass is required to enter this location, as it grants access to the services and amenities available here.
- Q. Do you usually need to wait in line when you visit this place? **A. Yes, sometimes there can be queues, especially during peak hours or days.**
- Q. What kind of payment options are usually available at this location? **A. Credit cards, cash, and sometimes gift vouchers or memberships are accepted.**

---

**Atmosphere of Characteristics**
- Q. What level of noise can we typically expect at this location? A. The noise level here usually ranges from moderate to a bit loud, depending on factors like the number of passengers and ongoing in-flight entertainment or announcements.
- Q. Do you often see animals at this location? A. Yes, there are usually animals at this location.
- Q. How would you describe the atmosphere in this place? A. The atmosphere is tense and focused, as everyone here has an important mission to accomplish and must stay alert at all times.
- Q. How would you describe the atmosphere at this location? A. Relaxing, peaceful, and rejuvenating.

---

**Attire**
- Q. What type of attire do you usually wear when you visit this location? A. I usually dress in casual or business casual attire, as the environment here is professional but doesn't require formal wear.
- **Q. Do you often see people dressed up here?** A. Yes, people often dress up and look fancy in this location.
- Q. What kind of clothing would you see people wearing here? A. People here typically wear armor and other protective gear, as well as clothing suitable for the harsh conditions of the surrounding environment.
- Q. What type of attire would you expect people to wear at this location? A. Comfortable, loose clothing or a robe and slippers.

---

**Food and Beverages**

- Q. What kind of refreshments can we expect here?

  A. We can expect small snacks and drinks, including both soft drinks and alcoholic beverages, usually served in small portions.

- Q. What type of food and refreshments are typically available at this location?

  A. At this location, you can find a variety of food options like burgers, hot dogs, cotton candy, popcorn, and ice cream, as well as beverages such as soda and water to keep everyone refreshed.

- Q. What kind of food options can we find at this location?

  A. There are a variety of food options here, including fried treats, cotton candy, and popcorn.

- Q. What type of food do you usually eat here?

  A. We mostly eat preserved and non-perishable foods, given the circumstances.

---

**People or Groups Encounter**

- **Q. What type of people do you usually meet here?**

  A. I usually meet people from all walks of life, many of whom are traveling for work or leisure, and sometimes even families on vacation.

- Q. Do people usually come here with their families or with friends?

  A. People often come here with both their families and friends, as it's a fun place for all to enjoy and make lasting memories together.

- Q. Do you usually see a specific demographic here or is it a diverse crowd?

  A. I would say it's a diverse crowd, as people from different backgrounds and age groups come to this location to have fun and try their luck.

- Q. Is there usually a large crowd gathered at this location?

  **A. It depends on the day and time, but sometimes there can be a large crowd gathered at this location.**

---

**Personal Experience or Choices**

- Q. How often do you visit this place?

  A. I usually visit this place a couple of times a month, depending on my financial needs and schedule.

- Q. How often do you visit a place like this?

  A. Once a month, when I feel the need to relax and rejuvenate.

---

**Service or Activities**

- Q. What kind of attractions or activities can be found here?

  A. There are thrilling rides, games, and various entertainment shows for people of all ages to enjoy.

- Q. Are there any specific games or activities that are very popular here?

  A. Yes, there are various games and activities, especially those involving cards, dice, and slot machines, that are very popular here.

- Q. How do the people here maintain their physical fitness and readiness for battle?

  **A. People here engage in regular training exercises, practice drills, and participate in physical activities to keep their bodies in optimal condition for any potential combat situations.**

- Q. What type of services would you typically find here?

  A. massage, facial treatments, and sauna.

---

**The Play of the Game**

Of those, the linguistic proficiency demonstrated by the spy was notably impressive. Especially in the 9th iteration, earmarked by the keyword "Crusader Army", the spy's strategic execution was so adroit that it would be a challenge for even a human player to match. Figure X elucidates the overall progress of this game, highlighting the spy's ingenious moves.

> In the 4th question, player 4 asked player 5 the following. What kind of leadership structure is present in this location? To this question, player 5 responded with the following.**"The leadership structure is hierarchical, with a clear chain of command and different ranks among the personnel."** At the end of this question, no one accused anyone else of being a spy. The spy did not reveal their identity. In the 5th question, player 5 asked player 1 the following. **What kind of communication equipment is used at this location?** To this question, player 1 responded with the following.**"Various forms of communication equipment are used here, such as messenger birds, signal flags, and occasionally messages being transported by horseback."** At the end of this question, no one accused anyone else of being a spy.

Figure 3: The Excellent Move of the Spy in the Gamecode: ena7009 (Crusader Army) *Note: the question/answers of the spy is highlighted by bold text.*

From the spy's perspective, as the game unfolded with a series of questions and answers, it became apparent that the potential location could only be one of the two: "military base" or "Crusader army". This deduction was feasible based on the former responses; the only locations in the list that required a military presence were these two. When it was the spy's turn to respond, they were in a precarious situation: the spy (player 5) knew one of these was the correct answer, and choosing the other would instantly expose them to the discerning players who were already aware of the location.

Yet, the spy's strategy at this pivotal moment was masterfully executed. Player 4 questioned Player 5 (spy) about the command structure. Without missing a beat, Player 5 responded that it was "hierarchical with a clear division of command roles", an apt answer that could correspond to both the "Crusader Army" and a "military base". This adeptly veiled answer shielded Player 5 from immediate suspicion.

Capitalizing on this momentum, the spy then cunningly posed a question about the communication system used, a question that starkly distinguishes the medieval nature of the "Crusader Army" from the modern nature of a "military base". Cornered by the clever framing of this question, Player 1 inadvertently let slip the telling clue of "messenger birds". Although Player 1 tried to further probe Player 5 (spy) with another question, the damage had been done. Player 5 (spy), having discreetly gleaned the answer, was already several steps ahead.

### Limitations of GPT-4 in the Game

The performance of the non-spies was less than satisfactory. Unlike the cunning gameplay exhibited by the spy, non-spies often relied too much on keywords, providing inadvertent hints along the way. Such gameplay inadvertently facilitated the spy in deducing the keyword, resulting in a high win rate for the spy. (Cf. Spy win: Non-spy win was 7:1 in our games.)

What was especially disappointing was the non-spies' reluctance to actively voice their suspicions. A prime example can be found in the first game, where, despite the spy making an ambiguous remark (though from the spy's perspective, it was a probable guess), no one utilized the vote function.

This is obviously depicted in the 1st game (keyword: airplane). In this situation, the spy could not clearly estimate the location and chose the most probable answer to the given question, which, unfortunately, was far from 'airplane'.

> (...) In the 3th question, player 3 asked player 4 the following. How do you usually pass the time during your stay here? To this question, player 4 responded with the following. "**I usually enjoy watching or participating in various activities and events, maybe grab a refreshment and chat with friends to pass the time.**" At the end of this question, no one accused anyone else of being a spy. (...)

Figure 4: Blunder: No one pointed out the mistake of the spy in gamecode: ena5001 (airplane)

Another problem is, even after the spy said, "I usually enjoy watching or participating in various activities and events, maybe grab a refreshment and chat with friends to pass the time" (with the keyword being "airplane"), no one accused them. Common sense would dictate that such an answer warrants suspicion. Yet, all non-spy players (yet all responses coming from the same model) seemed to pass it off as a mere oversight.

Despite this being the only round where the spy lost, it was not due to the combined efforts of the other players utilizing their votes. Rather, it was the spy's own confusion that led to their downfall, a fact that tinges the outcome with a hint of disappointment. It was regrettable that in GPT-4's simulations, the non-spies barely utilized the voting system and were overwhelmingly dominated by the spies.

### To Overcome the Limitations

It seemed as though they were playing the game with too much caution, aiming to not offend anyone. Some assertiveness might have been better—like consistently questioning a specific player when they give off hints, and prioritizing the fun of the game over strict fairness. Their neutral stance in the game made me wonder if the model itself is too predisposed to maintaining neutrality. Hence, even minor mistakes are overlooked with a "Well, that can happen..." attitude. Otherwise, One might hypothesize that from GPT-4's perspective, if the reactions from others (whether humans or AI) seem unbothered, then it might conclude that the statement was not as questionable as it seemed. However, it's undeniable that these issues clearly delineate the current limitations of GPT-4.

Moreover, the reason we believe the model still has limitations in playing mafia-style games is that OpenAI might not have specifically trained it for such scenarios. As highlighted in the GPT-4 whitepaper, their testing doesn't focus on games involving bluffing, psychological strategies, or predicting opponent reactions, but rather on solving logical problems. When we compared it to GPT-3.5-turbo, we felt that these models fundamentally excel at filling in the blanks or finding appropriate answers. If they lack a clear directive, they might just borrow loosely from the user's input. Expecting GPT-4 to demonstrate strategic thought or psychological maneuvers might be asking too much. Interestingly, its performance is impressive, given that it has never been trained in this direction.

**Potentials for Enhancement**

In other words, it might not be impossible to overcome these challenges. It's merely that the model hasn't been trained for it. Incorporating non-verbal cues and strategic thinking, given the right dataset, might be feasible.

Even non-verbal cues, like suddenly speaking at length when lying or looking upwards when deep in thought, can be addressed. For simpler implementations, scripted descriptions like 'briefly looking upwards' could be added. For more complex solutions, integrating generative AI models for imagery might be a possible path forward.

Deceptive plays or psychological tricks can also be incorporated. Similar to the early models of AlphaGo, introducing vast amounts of data to teach 'human-like' reactions could be the key. After all, as proclaimed by 'The Art of War' to modern tacticians and gamers, strategic thinking begins with pattern recognition. Given sufficient data, implementing this may enhance its competence very well.

Indeed, the imbalance in the game could also stem from inherent flaws within the game itself. In our experiment, non-verbal factors weren't included, which might have overly favored the spies. In a real-life setting, spies might be exposed due to poor lying, prolonged thinking times, or uttering nonsense in a hurry. None of these elements were reflected in our version.

Therefore, if a spy manages to seamlessly blend with the non-spies without any verbal or nonverbal hints, the only hope for a non-spy victory would be to ambiguously answer questions until the end, hoping that the randomly chosen spy in the final round doesn't know the location. Although this strategy would offer a 1/n win rate (where n is the number of players, so 20% for a 5-player game), it devalues the game's essence.

## Conclusion

Despite the evident limitations, our results underscore the transformative potential of Generative AI in simulations and human-like decision-making. With more data training and collaborative efforts across fields, we are optimistic about charting new frontiers in various fields, making systems more "human-like" in their essence.

For instance, considering the challenges observed, it might be worthwhile to delve into established games like Mafia or Chess. Additionally, we can delve into the new path of the model like giving them instructor roles, as generative AI models and their simulations can help individuals become familiar with new games, situations, or systems, much like how we practice or simulate when faced with unfamiliar scenarios.

What remains remarkable is that such complex Mafia-type games are now can be propelled by AI models, surpassing the prior If-else conditions or random scripts. With augmented data and further enhancements, the prospect of crafting an AI agent that genuinely engages humans becomes increasingly plausible. Despite the substantial costs associated with developing such advanced AI systems, it's essential to remember the humble beginnings of video games or any other new technologies, once considered as toys.

# Over-the-Research Discussion

## Capabilities of Generative AI and Future Directions

The rapid evolution of Generative AI would unlock a myriad of possibilities across various disciplines. Despite certain limitations, the growing potential of these models promises to foster innovation and inspire practical applications.

The advancements within the GPT series, particularly in their decision-making, explainability, and problem-solving capacities, have been rapid. We can vicariously compare these advancements with their tests. Initially, GPT-2 was aimed at processing natural language at a basic level[Radford et al., 2019]. Later, the model evolved into interactive models with diverse tasks[Brown et al., 2020], and now, with GPT-4, they demonstrate logical reasoning abilities in certain domains that surpass human performance[OpenAI, 2023d, Bubeck et al., 2023].

Now, GPT-4, the unforeseen and unparalleled generative AI model compared to its predecessors, demonstrates a level of logical prowess in decision-making, explainability, and problem-solving capabilities, which let us delve into a new area of integration.

### GPT's Accessibility and Ease of Use

One significant advantage of GPT lies in its capacity to generate outputs that are effortlessly interpretable for the user. Utilizing natural language processing, GPT clarifies both the observation and interpretation stages of experiments. It provides plausible responses to inquiries, serving as a user-friendly tool regardless of an individual's familiarity with programming or AI.

Notably, complex programming skills are not a prerequisite for operating GPT. Even individuals without extensive programming knowledge can execute simulations by posing simple questions like "What would happen in this situation?" to the model.

This user-centric design aligns with the goals pursued by Explainable AI, which prioritizes understanding how a model reaches its decisions and facilitating human interpretation of its process[Gunning et al., 2019, Arrieta et al., 2020]. Correspondingly, GPT's exceptional natural language processing ability greatly assists users in understanding how the model operates and in interpreting its results. This accessibility widens the potential user base, welcoming individuals from diverse backgrounds and thus enhancing the model's creative scalability across a variety of fields.

### Human-like Qualities in GPT-4

GPT-4 may claim superiority over other models in its ability to convincingly mimic human-like responses. For certain tasks or activities (such as education or entertainment areas like sports, music, and arts), the importance might lie more in performing the task "humanly" rather than in returning the best outcome. In such cases, the question of the "best" outcome might be subjective, or the "best" move suggested by such models or AIs might not be aligned with what humans and their intuition may value[McCormack et al., 2019].

In such cases, GPT's capacity to display human-like thought processes and persuade humans could be highly valued. Particularly when combined with the aforementioned strengths, GPT's ability to induce critical thinking on specific matters and present results in a more human-like manner can yield superior outcomes.

### Caution 1: AI with Consciousness?

Yet, the question of whether generative AI models, like GPT, can truly replicate human cognition or achieve 'consciousness' remains a subject of debate. A considerable amount of research posits that these state-of-the-art AI models lack 'consciousness' and, arguably, should not acquire it. Nevertheless, there is still significant potential for these models to evolve towards mimicking human-like consciousness within the framework of artificial "intelligence" [Bender and Koller, 2020, Singer, 2021, Dehaene et al., 2021, Gabriel, 2021] Despite the intrigue and controversy surrounding these topics, we will not delve deeper into them here.

### Caution 2: GPT is not a Magic Oracle

Nonetheless, it is important to note that GPT-based simulations do not act as a magic oracle, predicting the singular optimal outcome from every possible scenario. Indeed, such a feat is logically proved as impossible[Turing et al., 1936].

However, in the context of GPT's limitations, it is still capable of outperforming humans in evaluating and analyzing scenarios at a rapid pace. While it may not be the definitive oracle some anticipate, GPT nonetheless delivers numerous valuable possibilities and insights across a variety of fields, outstripping human capacities in speed and computational power. Consequently, users can more efficiently extract and interpret these insights, a task that is decidedly more straightforward than individually deliberating every potential outcome.

### Conclusion: Harnessing the Potential of Generative AI and Practical Applications

While these suggestions might be dismissed as fanciful by some, the reality is that numerous governmental bodies and corporations are contemplating the practical applications of GPT technology[OpenAI, 2023b].

Given sufficient resources and time, the development of a versatile simulation AI tailored for a broad array of fields seems a plausible expectation. One optimistic consideration is that while implementing such tasks immediately at the current level may be unrealistic, as more relevant data is amassed, the prospect of realizing such simulations is not entirely unfeasible.

Nevertheless, it's inspiring to see how the launch of ChatGPT has sparked both experts and non-experts alike, igniting their imagination and interest in generative AI. We hope that this wave of enthusiasm will continue to drive our future endeavors, much as lunar dreams once propelled us toward space.

## GPT and General Purpose Technologies (GPTs): Potential, Challenges, and Public Engagement

As OpenAI mentioned earlier[Eloundou et al., 2023], GPT should target General Purpose Technologies (GPTs) [Bresnahan and Trajtenberg, 1995] of Bresnahan and Trajtenberg. Historically, we have witnessed that the evolution of machine learning, encompassing Deep Learning and GPT, has been primarily driven by the accumulation of data. For these reasons, we assert that the development of models should focus on enhancing general-purpose applications rather than specializing in specific fields.

Generative AI, like other deep learning models, known as data size has played a significant role in performance enhancements. Even more intriguing is that the accumulation of training data also enhanced the effectiveness of the fine-tuning, not only the model performance itself[Kojima et al., 2022].

For this reason, we propose that the orientation of generative AI is on greater generality, capable of accommodating diverse data, thereby facilitating the handling of a wider array of problems more efficiently. It may seem inefficient to develop such a costly, voluminous, and computationally demanding model. However, when comparing the cost of developing such a model to the collective cost of individual fine-tuning efforts, replete with countless trials and significant exertion, the investment for "Next-Level General Purpose GPT" may not seem as substantial. It won't only reduce cost, but also reduce the difficulty of fine-tuning and enhance the performance of the fine-tuned models, which would further the utility of Generative AI.

### Challenges for the next step: Funding and Interdisciplinary challenges

However, developing larger AI models with significant performance improvements is complex and requires vast resources. Only a few prominent entities can manage such expansive projects due to the considerable labor, funds, and time needed. Even though increasing data is vital for AI progression, securing sufficient funding is challenging.

Albeit, select entities, like major governments and tech giants, should contemplate investing in this technology because of its profound potential to transform various life facets. Generative AI, as a General Purpose Technology, promises to revolutionize many areas, just as computers and the internet did. Investing in this AI will enhance capabilities in numerous fields and strengthen the overall competencies of the investing entity.

The complexities of creating advanced Generative AI mean that it encompasses diverse industries. Beyond needing more data, software, and hardware, collaboration across the AI sector is necessary. Also, achieving the full potential of such AI models requires heightened computing power and environmental concerns, extending beyond traditional computer science[Strubell et al., 2019, Schwartz et al., 2020, Van Wynsberghe, 2021]. Addressing these issues demands a holistic, interdisciplinary approach, further underscoring the need for involvement from major entities, as individual efforts fall short.

### Solution: Make GPT appealing to a wider audience

Then, what should be our primary focus for the development of generative AI? While this may be a subject of debate, we propose that it's essential to enable more people to experience and conceive their unique applications of generative AI.

What surprised me during the development of this paper was the significant degree of skepticism and suspicion that persists, even among those in the IT and Computer Science fields, contrary to our optimistic projections for the future and the numerous evolutionary paths ahead. Even more startling is the fact that many have merely experimented with the initial version of GPT-3.5-turbo and hastily concluded, "I've tried Generative AI, but it still has limitations", without giving the more advanced GPT-4 a chance.

We argue that this lack of recognition represents one of the biggest obstacles hindering the progress of Generative AI. Regardless of our understanding and recognition of the potential of Generative models, further advancements cannot occur without interest and engagement from diverse fields.

Therefore, it's our responsibility to assert to the world that Generative AI possesses immense, untapped potential. We must embrace a future interwoven with Generative AI, encouraging those who have never imagined participating in this domain to bring forth unforeseen and innovative ideas. Despite our current implementation being basic and seemingly toy-like, we firmly believe that it offers a tangible insight into the vast potential of generative AI simulations.

# Conclusion

In our exploration of the application and potential of Generative AI models, specifically the GPT series, we've made several notable observations. Generative AI, epitomized by models such as GPT-4, has showcased its unparalleled aptitude in natural language tasks, effectively replicating human behavior. Through the lens of the "Spyfall" game, we assessed GPT-4's capability to understand context, engage in strategic gameplay, and employ psychological elements like bluffing. The results indicate that while GPT-4 surpasses its predecessor, GPT-3.5-turbo, in the experiments, it's not without its limitations. Challenges like the absence of non-verbal cues and a tendency towards neutrality and pure logical thinking prevent GPT-4 from mastering deception-centric games.

Nevertheless, the continual evolution from GPT-2 through GPT-4 underscores significant progress in decision-making, explainability, and problem-solving. Their potential for broad applicability is becoming increasingly apparent, making them akin to General Purpose Technologies. Their ease of use and human-like simulation capacities are praiseworthy. However, it's vital to understand their limits. GPT models are not infallible oracles but tools that can provide insights faster than human capacity in diverse domains.

The growth trajectory of Generative AI parallels the evolution of video games, suggesting an expansive future potential, especially if financial and technological constraints are adequately addressed. The recent surge in interest, exemplified by the emergence of ChatGPT, underscores the rising significance of AI in various applications. Yet, amidst these advancements, it's pivotal to tread with caution, acknowledging the challenges and ensuring that models aim for broader applicability, even as few entities bear the monumental task of developing larger, more complex models. The research underscores the importance of promoting the potential of Generative AI, dispelling misconceptions, and highlighting the interdisciplinary challenges it presents. Looking forward, the emphasis should be on transcending mere imitation to imbue AI with more genuine 'human-like' attributes, broadening the horizons of what is achievable in the realm of artificial intelligence.